\documentclass[10pt,twocolumn]{article}
\usepackage[letterpaper,margin=0.75in,columnsep=0.28in]{geometry}
\usepackage[T1]{fontenc}
\usepackage{microtype}

\usepackage{cite}
\usepackage{enumitem}
\usepackage{amsmath,amssymb,amsfonts}
\usepackage{algorithm}
\usepackage{algorithmic}
\usepackage{graphicx}
\usepackage{textcomp}
\usepackage{xcolor}
\usepackage{url}
\usepackage{tikz}
\usepackage{pgfplots}
\pgfplotsset{compat=1.17}
\usepackage{booktabs}
\usepackage{threeparttable}
\usetikzlibrary{shapes.geometric,arrows.meta,positioning,fit,backgrounds,calc}

\definecolor{measuredblue}{RGB}{70,130,180}
\definecolor{inferredorange}{RGB}{235,125,20}
\definecolor{aigreen}{RGB}{50,150,50}
\definecolor{gapgray}{RGB}{150,150,150}

\makeatletter
\renewcommand{\@maketitle}{%
  \null\vskip 0.3em%
  \begin{center}%
    {\rule{\textwidth}{0.6pt}}\\[0.95em]%
    {\Large\bfseries \@title\par}%
    \vskip 0.75em%
    {\rule{\textwidth}{0.6pt}}\\[1.15em]%
    {\large \@author\par}%
  \end{center}%
  \vskip 1.2em}
\makeatother

\begin{document}

\title{Activity Frames: Deterministic Screen-Activity Compilation\\ for Agent Memory and Replay}

\author{Nossa Iyamu\thanks{Independent Researcher. Correspondence to \texttt{nossa.iyamu1@gmail.com}.}}
\date{}

\maketitle

{\centering\bfseries\large Abstract\par}
\smallskip
\noindent
Computer-use agents pay full frontier inference to re-derive routines their user has already performed, because an agent's memory today records what the user said, not what the user did. We compile passively captured screen activity into agent memory with a deterministic, zero-model pipeline: it segments a local capture stream into typed \emph{activity frames}, bounded episodes carrying application, site, timing, input volume, and evidence pointers back to the raw rows, with no model in the loop, so the output is byte-identical, cacheable, and mechanically auditable. On one professional's single-user corpus of $128{,}756$ frames over $51$ active days, the compiler reduces a day of raw capture to a prompt-ready context block $86\times$ smaller in $68$\,ms, and an agent reading that block answers questions about the day at $98.4\%$ accuracy (Wilson $95\%$ CI $91.7$--$99.7\%$) against an independent oracle, versus $66$--$80\%$ for an LLM summary of the same capture, a mid-tier model reading the block matching a frontier one.

The same compiler doubles as a demand-side cost instrument. Read off passive, pre-delegation human activity rather than agent rollouts, it supplies two parameters that agent-cost models assume but, to our knowledge, have not measured: the Routine Overhead Ratio $R$ and the routine recurrence $h$. We report first values of $R$, a modeled upper bound, at $60$--$343\times$, and a delegable recurrence of $9.0\%$ in-sample and $7.7\%$ out-of-sample, for a realistic all-fleet token ceiling near $8\%$; a compiled routine replays deterministically with the model out of the loop, demonstrated live at zero model tokens on a guard-matched hit. Schema, compiler, and evaluation harness are open.
\medskip

\medskip

\section{Introduction}
An LLM agent asked ``what should I prioritize today?'' answers from what it can see: the conversation, some files, perhaps a calendar. It cannot see that its user spent the morning inside a pull request, switched contexts forty times after lunch, or abandoned a draft email at 16:40. Position work has argued that episodic memory, a record of experience situated in time, is the missing piece for long-term LLM agents \cite{pink2025position}, and surveys of agent externalization treat memory as a first-class architectural concern \cite{zhou2026externalization}. Yet in practice, agent memory today means conversation memory: what the user told the model, not what the user did \cite{packer2023memgpt, chhikara2025mem0, rasmussen2025zep}.

The raw material for behavioral episodic memory already exists. Automated time trackers have recorded application focus for a decade \cite{activitywatch}; operating systems now ship continuous capture, with Microsoft Recall storing periodic snapshots on Copilot+ PCs \cite{recall2025}; and OpenAI's Chronicle preview builds memories from recent screen content for its Codex assistant \cite{chronicle2026}. Capture, in other words, is commodity. Consumption is not. A day of event-driven capture on our corpus yields roughly two thousand snapshot rows, each one asserting only that at time $t$, application $a$ displayed window $w$ at URL $u$. Two consumption strategies dominate. Raw search hands the agent a flat result list and leaves sessionization, deduplication, and duration accounting to the model at inference time; we measure this cost at 126{,}812 tokens for a single day (Section~\ref{sec:eval}). LLM summarization compresses well but imports the weaknesses of its summarizer: per-run cost, non-determinism, degraded reliability over long inputs \cite{liu2023lost}, and the possibility of inventing activity that never happened. Neither yields memory an agent can cache, audit, or trust.

This paper proposes the missing middle: \emph{deterministic compilation}. In the same way a compiler turns instructions into structured artifacts without opinion, we turn snapshot streams into \emph{activity frames}: bounded episodes with an application and site, a start and end, dwell-based active time, typed page references, input volume, and evidence pointers back to the raw rows. No model participates. The same database and window always produce the same document, so episodic memory becomes free to rebuild, safe to cache, and mechanically auditable. Interpretation is not banned but quarantined: a second schema tier accepts inferred labels only when they are namespaced, confidence-tagged, and linked to the measured evidence that supports them.

Our contributions are:
\begin{itemize}[leftmargin=*]
    \item \textbf{A two-tier schema} for representing human computer activity to agents, separating measured fact from tagged inference, with coverage gaps and blind spots as mandatory document elements (Section~\ref{sec:schema}).
    \item \textbf{Deterministic compilation rules} (dwell crediting, session gaps, flicker merging, nearest-frame attribution, coordinate-based click resolution, total URL-to-entity typing) that require no learned components (Section~\ref{sec:compile}).
    \item \textbf{An open reference implementation}: a dependency-free Python compiler with a provisioned local capture engine, a CLI, and a Model Context Protocol (MCP) server exposing six tools to any MCP-capable agent \cite{activityframes2026, mcp2024} (Section~\ref{sec:impl}).
    \item \textbf{An empirical characterization} on a 61-day live corpus: token cost against raw rows (86$\times$ reduction for prompt-ready context), 68\,ms full-day compilation, byte-identical reproducibility, entity-typing coverage, and a duration distribution that separates genuine short attention bouts from single-snapshot transits and multi-monitor effects (Section~\ref{sec:eval}).
    \item \textbf{A downstream question-answering benchmark} against an independent oracle, run at two model tiers, showing that an agent answers more accurately from the compiled block than from raw rows or an LLM summary of the same capture at both tiers, that the block lets a mid-tier model match a frontier one, and that a stronger model narrows but does not close the gap; the harness is released for rerunning against any model (Section~\ref{sec:qa}).
    \item \textbf{A demand-side cost instrument}: an open, deterministic measurement of the Routine Overhead Ratio $R$ (modeled numerator, measured denominator) and desktop routine recurrence $h$, read from passively captured human activity rather than agent rollouts, with a parametric replay executor and a first live proof-of-concept, reported on one user's corpus (Sections~\ref{sec:ror}--\ref{sec:recurrence}).
\end{itemize}

\section{Related Work}\label{sec:related}

\subsection{Memory Systems for LLM Agents}
MemGPT virtualizes context OS-style, paging conversation history through a bounded window \cite{packer2023memgpt}; production layers consolidate salient facts from dialogue \cite{chhikara2025mem0}; Zep organizes memory as a temporal knowledge graph \cite{rasmussen2025zep}; E-mem reconstructs episodic context from execution traces \cite{wang2026e}; and hierarchical procedural memory distills skills from agent trajectories \cite{forouzandeh2025learning}, threads a recent review unifies as externalization \cite{zhou2026externalization}. All ingest what flowed \emph{through the agent}, messages, tool calls, task traces, and memory-construction studies show that granularity and structure affect retrieval quality \cite{pan2025on}, supporting typed episodes over flat logs. Agent Workflow Memory induces routines too, but from agent rollouts rather than human activity \cite{wang2024agent}.

\subsection{Screen Capture as Agent Memory}
A 2025--2026 wave targets our exact goal by the opposite method. MIRIX runs a multi-agent memory system over continuous screenshots with cloud model calls \cite{wang2025mirix}; FOCAL invokes a local vision-language model to write session summaries from on-device capture, reporting a 60\% token cut \cite{yin2026focal}; ProAgentBench segments 500+ hours of capture on application switches before model-based annotation \cite{tang2026proagentbench}; SummAct summarizes interaction traces into intentions with an LLM \cite{zhang2024summact}; and OmniQuery augments captured personal media for QA \cite{li2024omniquery}. In every case a model sits inside the memory-construction loop, so the memory inherits per-run cost, non-determinism, and possible hallucinated episodes. Activity frames are the deterministic counterpoint: no model in the compile path, byte-identical output, and interpretation quarantined in a separate evidence-linked tier, the first deterministic desktop instantiation of the acquisition layer episodic-memory advocates call missing \cite{pink2025position}.

\subsection{Desktop Activity, GUI Agents, and Memory Trust}
Deriving task structure from interaction logs is a two-decade ambition: TaskTracer tied window, file, and clipboard events to declared tasks \cite{dragunov2005tasktracer}, and SWISH clustered windows into tasks from titles and switching \cite{oliver2006swish}. Interruption science established the fragmentation our compiler measures, three-minute working spheres \cite{gonzalez2004constant} and content switching about every 19\,s with 75\% of segments under a minute \cite{yeykelis2014multitasking}; these are behavioral rates from their instruments, and our comparable figure is the post-transit median frame of $0.9$\,min (Section~\ref{sec:eval}), not our $6.1$\,s median inter-capture gap, which is recorder cadence rather than a human switching rate. Methodologically our pipeline descends from lifelogging's threshold segmentation \cite{doherty2008segmenting}, whose critique that unstructured total capture serves nobody \cite{sellen2010beyond} is the failure mode we exist to prevent, plus web sessionization's inactivity timeouts \cite{cooley1999data}, process-mining event abstraction \cite{vanzelst2021event}, robotic process mining's UI-log routine identification \cite{leno2021robotic}, and the data-to-text faithfulness tradition \cite{portet2009babytalk}; but those pipelines end in process models or prose for humans, ours in typed memory for an agent. Separately, a large body gives agents screens as an \emph{action} surface, GUI-agent surveys cataloging accessibility-tree, OCR, and vision perception \cite{zhang2024large, liu2025llm}, and generative agents showing a remembered observation stream enables long-horizon behavior \cite{park2023generative}; we compile the inverse, what the human did, an acquisition problem personal-agent surveys call open \cite{li2024personal}. Deployed capture leaves the same gap: ActivityWatch is content-blind \cite{activitywatch}, Recall exposes snapshots to the user but no agent API \cite{recall2025}, Chronicle feeds one walled assistant \cite{chronicle2026}. Finally, agent memory is now an attack surface, memories poisoned to steer behavior \cite{louck2026securing} or probed by membership inference \cite{chen2026mrmmia}, motivating two properties we make structural: evidence pointers back to raw rows, and a hard measured/inferred boundary, so a poisoned or hallucinated label cannot pass as fact. Context-engineering and long-context degradation results both argue for compact structured context over raw dumps \cite{mei2025a, liu2023lost}.

\subsection{Agent Skills, Trajectories, and Cost}\label{sec:rel-cost}
Three adjacent lines each need one of our parameters but source it differently. \emph{Skill induction} turns an agent's own successful trajectories into reusable skills: Agent Workflow Memory from web rollouts \cite{wang2024agent}, SkillWeaver from self-execution \cite{skillweaver2025}, Agent Skill Induction as programs \cite{asi2025}, and PreAct as continual self-improvement \cite{preact2026}. All read the \emph{supply} side, what an agent did while acting, and so observe a routine only after an agent has already performed it at full cost, and only for tasks it can complete. \emph{Trajectory-acquisition} channels manufacture training data and report a price: Explorer at ${\approx}\$0.28$ each \cite{explorer2025}, AgentTrek at ${\approx}\$0.55$ \cite{agenttrek2024}, and Watch\,\&\,Learn \cite{watchlearn2025} and cotomi Act \cite{cotomiact2026} at low marginal cost; but a synthesized trajectory exists only where the generator can drive the task to success, so authenticated, private-state routines are systematically under-covered, and Agent Data Protocol's channel taxonomy \cite{adp2025} omits both passive capture and a price column. \emph{Cost frameworks} price the rest: FrugalGPT for model cascades \cite{chen2023frugalgpt}, Cost-of-Pass for expected cost per success \cite{erol2025costofpass}, AI Agents That Matter for cost as a first-class axis \cite{kapoor2024agentsmatter}, and the Holistic Agent Leaderboard for real rollout economics (${\approx}\$1.84$/task) \cite{hal2025}. Together they supply an amortized per-task cost of a memory-backed agent, which we adopt as a cited frame, not a contribution:
\begin{equation}
\begin{split}
\mathbb{E}[\$/\text{task}] = {}& (1{-}hq)\tfrac{C_\text{miss}}{p} + hq\,(C_\text{hit}{+}C_\text{verify}) \\
&{}+ h(1{-}q)\,C_\text{wrong} + \tfrac{C_\text{write}}{N},
\end{split}
\label{eq:amortized}
\end{equation}
where $h$ is recurrence, $q$ the fraction of hits correctly matched, $p$ the base success rate, $N$ the reuse count, and the $C$-terms per-branch costs. Every term is already priced in the works above; the one input none of them measures is $h$ on the \emph{pre-delegation} passive corpus. Rollout-priced leaderboards observe the hit rate of tasks an agent was given and completed, a survivorship-biased quantity; the demand-side $h$, how often the user's real work repeats whether or not an agent could do it, is what our instrument supplies, and $R$ replaces the assumed $C_\text{miss}/C_\text{hit}$ ratio with a measured one. Activity frames are thus the pre-delegation, demand-side instrument that skill induction, acquisition channels, and cost leaderboards each presuppose but none provides.

\section{The Activity Frames Schema}\label{sec:schema}

\subsection{Design Principles}
Four principles fix the schema's character. \textbf{Measured, not guessed}: every field in a standard document is derivable by deterministic code from capture data; there are no intent labels, because code cannot observe intent. \textbf{Reproducible}: identical inputs must yield identical documents, making memory cacheable and diffable. \textbf{Evidenced}: every frame carries pointers to the raw rows it was compiled from. \textbf{Honest about absence}: periods without capture are reported as gaps, and every document carries a \texttt{blind\_spots} list stating what the pipeline systematically cannot see. Consumers must treat uncovered time as unknown, never as inactivity.

\begin{figure}[t]
    \centering
    \begin{tikzpicture}[
        node distance=0.55cm and 0.55cm,
        box/.style={rectangle, rounded corners=2pt, draw, fill=#1!12, text=black,
            minimum width=2.35cm, minimum height=0.95cm, align=center,
            font=\sffamily\scriptsize},
        lbl/.style={font=\sffamily\scriptsize\bfseries, text=#1!60!black},
        arr/.style={-{Stealth[length=2mm]}, thick}
    ]
        \node[box=gapgray] (cap) {capture engine\\ \tiny screen, a11y tree,\\ \tiny input events (SQLite)};
        \node[box=measuredblue, right=of cap] (comp) {compiler\\ \tiny sessionize, enrich,\\ \tiny entity typing};
        \node[box=aigreen, right=of comp] (agent) {agent\\ \tiny MCP tools,\\ \tiny context blocks};
        \node[box=inferredorange, above=0.45cm of agent, dashed] (inf) {tier 2: inferred\\ \tiny namespaced, confidence-\\ \tiny tagged, evidence-linked};
        \draw[arr] (cap) -- node[above, font=\tiny]{read-only} (comp);
        \draw[arr] (comp) -- node[above, font=\tiny]{tier 1} (agent);
        \draw[arr, dashed, inferredorange] (inf) -- (agent);
        \draw[arr, dashed, inferredorange] (comp.north) to[bend left=18] node[above left, font=\tiny, text=inferredorange!70!black]{optional} (inf.west);
        \begin{scope}[on background layer]
            \node[fit=(cap)(comp)(agent), draw=gapgray!60, dashed, rounded corners,
                  inner sep=5pt, label={[font=\tiny\sffamily, text=gapgray!80!black]below:{local machine; nothing leaves by default}}] {};
        \end{scope}
    \end{tikzpicture}
    \caption{The two-tier architecture. The measured tier (blue) is produced entirely by deterministic code reading the capture database. Interpretation (orange) is an optional extension that must be namespaced, confidence-tagged, and evidence-linked; stripping it always leaves a valid measured document.}
    \label{fig:arch}
\end{figure}
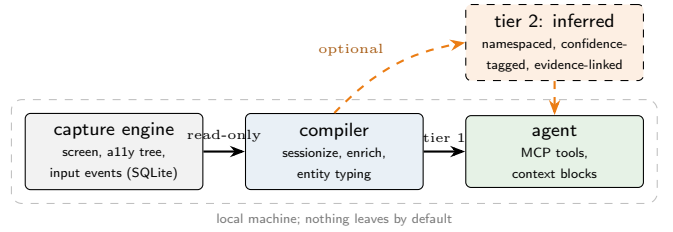

\subsection{Document Structure}
A document describes one query window and contains four parts: a \texttt{coverage} section (first and last activity, active minutes, span, capture gaps over five minutes), a chronological list of \texttt{frames}, a \texttt{blind\_spots} list, and provenance metadata (\texttt{schema\_version}, generation time, source recorder). A frame is one bounded stretch of attention in a single context, keyed by the pair (application, site), where site is the URL host for browser activity and absent otherwise. Figure~\ref{lst:example} shows a representative frame.

\begin{figure}[t]
\begin{center}
\footnotesize
\begin{tabular}{p{0.92\columnwidth}}
\toprule
\texttt{- id: f-0007}\\
\texttt{~~app: "Google Chrome"}\\
\texttt{~~site: "linkedin.com"}\\
\texttt{~~start: "20:24:04"~~end: "20:42:11"}\\
\texttt{~~duration\_min: 18.0~~~wall\_min: 21.5}\\
\texttt{~~pages:}\\
\texttt{~~~~- \{kind: people\_search, entity: "cto paris", count: 2\}}\\
\texttt{~~~~- \{kind: profile, entity: "jane-doe"\}}\\
\texttt{~~~~- \{kind: company, entity: "acme"\}}\\
\texttt{~~input: \{keys: 214, clicks: 31\}}\\
\texttt{~~interruptions: [\{app: "Slack", seconds: 12\}]}\\
\texttt{~~evidence: \{frame\_ids: "99871..100147"\}}\\
\bottomrule
\end{tabular}
\end{center}
\caption{A single activity frame (YAML, entities anonymized). Every field is computed by code; the evidence pointer names the raw snapshot rows the frame was compiled from.}
\label{lst:example}
\end{figure}

\subsection{Typed Page References}
Browser frames carry \texttt{pages}: typed references produced by deterministic URL parsing. A reference has a \texttt{kind} (what sort of page), an optional \texttt{entity} (the human-relevant identifier), and a view count. Standard kinds include \texttt{profile}, \texttt{company}, \texttt{people\_search}, \texttt{search}, \texttt{repo}, \texttt{pull\_request}, \texttt{issue}, \texttt{doc}, \texttt{email}, \texttt{video}, \texttt{post}, \texttt{ai\_chat}, and \texttt{local\_dev}. The mapping is total: a URL matched by no site parser falls back to a generic \texttt{page} reference with its domain, so typing never loses data. Kinds are open for extension but must remain deterministic functions of the URL.

\subsection{Tier 2: The Inferred Extension}
Tools may add interpretation, such as task labels or project clusters, under three schema-enforced rules: inferred content lives in a namespaced \texttt{inferred} block, carries a \texttt{confidence} tag drawn from \{\texttt{high}, \texttt{medium}, \texttt{speculative}\}, and names its \texttt{evidence}: the measured fields or raw rows supporting it. A consumer can always strip the \texttt{inferred} block and be left with a purely measured document. The reference implementation emits tier 1 only; we consider the boundary itself, not any particular inference method, to be the contribution.

\subsection{Privacy Rule}
Input \emph{volume} (keystroke, click, and copy counts) is part of the standard document. Input \emph{content} (typed text) must be excluded by default and included only on an explicit operator opt-in. This is a schema requirement, not an implementation courtesy: a conforming producer cannot silently emit content.

\section{Deterministic Compilation}\label{sec:compile}

\subsection{Setting}
The capture engine is event-driven with a heartbeat: it stores a snapshot row on interaction and on screen change (clicks, application switches, visual changes), plus a periodic row after roughly 30 seconds without input. Heartbeat rows are 19\% of our corpus; they matter because they keep the stream alive while the user reads, watches, or steps away from an awake display, and every downstream time measure inherits that. Each monitor records its own stream; within a stream the median inter-frame gap is 6.1\,s and the 90th percentile is 30.7\,s (Section~\ref{sec:eval}). A row carries a timestamp, application, window title, and URL when the focused application is a browser. Input events (keystrokes, clicks, clipboard, application switches) arrive in a parallel stream. Compilation consumes these streams for a query window, segmenting each monitor independently, and emits the document of Section~\ref{sec:schema}.

\subsection{Sessionization}
\begin{figure}[t]
    \centering
    \begin{tikzpicture}[xscale=0.69]
        \draw[-{Stealth}] (0,0) -- (11.6,0) node[below left, font=\tiny]{time};
        \foreach \x in {0.3,0.7,1.1,1.6,2.0,2.5} \draw[measuredblue, thick] (\x,0) -- (\x,0.28);
        \draw[measuredblue, |-|, thick] (0.3,0.75) -- (2.5,0.75) node[midway, above, font=\tiny, text=measuredblue!70!black]{A (linkedin.com)};
        \foreach \x in {2.8,3.0} \draw[inferredorange, thick] (\x,0) -- (\x,0.28);
        \draw[inferredorange, |-|, thick] (2.8,0.45) -- (3.0,0.45) node[midway, above, font=\tiny, text=inferredorange!80!black]{B 12s};
        \foreach \x in {3.3,3.8,4.3,4.9} \draw[measuredblue, thick] (\x,0) -- (\x,0.28);
        \draw[measuredblue, |-|, thick] (3.3,0.75) -- (4.9,0.75) node[midway, above, font=\tiny, text=measuredblue!70!black]{A cont.};
        \draw[measuredblue!60, dashed, thick] (2.5,0.75) -- (3.3,0.75);
        \node[font=\tiny, text=inferredorange!80!black] at (2.9,1.12) {flicker merged};
        \draw[gapgray, decorate, decoration={zigzag, segment length=5pt, amplitude=1.2pt}] (5.2,0.14) -- (7.6,0.14);
        \node[font=\tiny, text=gapgray!90!black] at (6.4,0.5) {gap 41 min (reported)};
        \foreach \x in {7.9,8.5,9.4,10.6} \draw[aigreen, thick] (\x,0) -- (\x,0.28);
        \draw[aigreen, |-|, thick] (7.9,0.75) -- (10.6,0.75) node[midway, above, font=\tiny, text=aigreen!60!black]{C (github.com)};
        \draw[{Stealth}-{Stealth}, thin] (9.4,-0.32) -- (10.6,-0.32);
        \node[font=\tiny, anchor=east] at (10.6,-0.6) {$\Delta t{=}104\mathrm{s} \Rightarrow$ credit $\min(\Delta t, 90\mathrm{s})$};
    \end{tikzpicture}
    \caption{Segmentation on one timeline. Ticks are snapshot rows. A 12-second detour ($B$) folds into the surrounding frame as a recorded interruption; a long silence becomes a reported coverage gap; sparse snapshots earn at most the 90\,s dwell cap.}
    \label{fig:timeline}
\end{figure}
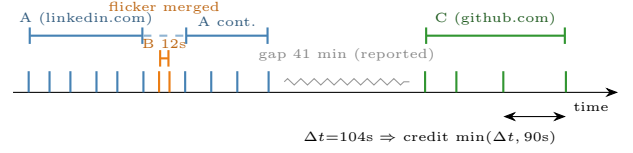

Three constants govern segmentation, chosen from the capture cadence rather than tuned on outcomes. \textbf{Dwell}: a frame contributes $\min(\Delta t, 90\,\mathrm{s})$ of active time, where $\Delta t$ is the gap to the next frame in its monitor's stream. Because the engine emits a $\sim$30\,s heartbeat during input-free stretches, dwell measures \emph{screen tenure}: how long a context stayed frontmost on an awake display. That includes reading and watching, which produce no input, but also stretches where the user has stepped away while the display stays on; Section~\ref{sec:limits} quantifies this. The cap, roughly three heartbeat periods, bounds the credit a single frame can earn when capture stalls or the display sleeps; it does not, and cannot, distinguish attention from presence. Consumers who need interaction-gated time can compare each frame's reported input volume against its duration. Segmentation runs per monitor, so a context visible on two monitors at once earns tenure on both; the schema discloses this as a blind spot and Section~\ref{sec:qa} shows the consequence. \textbf{Session gap}: a gap above 300\,s closes the current frame and becomes a candidate coverage gap; no dwell is credited across it. \textbf{Flicker merge}: the pattern $A \to B \to A$, where $B$ lasts at most 20\,s of wall time and no session gap intervenes, collapses into a single $A$ frame. Crucially, $B$ is not discarded: it is recorded on the merged frame as an \texttt{interruption} with its measured seconds, and its time is \emph{not} added to $A$'s active duration. Nothing is hidden and nothing is double-counted. Algorithm~\ref{alg:segment} summarizes the procedure; Figure~\ref{fig:timeline} illustrates all three rules on one timeline.

\begin{algorithm}[t]
\caption{Frame segmentation (one pass, then flicker merge)}
\label{alg:segment}
\begin{algorithmic}[1]
\REQUIRE snapshots $f_1 \dots f_n$ of one monitor's stream, sorted by time; constants $D{=}90$, $G{=}300$, $F{=}20$ (seconds)
\STATE $S \leftarrow []$;\quad $cur \leftarrow \bot$
\FOR{$i = 1$ \TO $n$}
    \STATE $k_i \leftarrow (\mathrm{app}(f_i), \mathrm{site}(f_i))$;\quad $\Delta \leftarrow t(f_{i+1}) - t(f_i)$
    \IF{$cur = \bot$ \OR $k_i \neq \mathrm{key}(cur)$}
        \STATE append new segment $cur$ with key $k_i$ to $S$
    \ENDIF
    \STATE extend $cur$ with $f_i$
    \IF{$\Delta \le G$} \STATE $\mathrm{active}(cur) \mathrel{+}= \min(\Delta, D)$
    \ELSE \STATE $cur \leftarrow \bot$ \COMMENT{session break; candidate gap}
    \ENDIF
\ENDFOR
\FOR{consecutive $A, B, A'$ in $S$ with $\mathrm{key}(A){=}\mathrm{key}(A')$}
    \IF{$\mathrm{wall}(B) \le F$ \AND no session break around $B$}
        \STATE merge $A'$ into $A$; record $B$ as interruption of $A$
    \ENDIF
\ENDFOR
\RETURN $S$
\end{algorithmic}
\end{algorithm}

\subsection{Enrichment}
Raw input events have three reliability defects that code can repair. First, \textbf{stale attribution}: the recorder sometimes tags an event with the previously focused application. Each event is therefore re-attributed to the temporally nearest snapshot (binary search over the frame stream), whose application, window, and URL are authoritative; the attribution distance is kept in milliseconds so downstream consumers can judge it. Second, \textbf{anonymous clicks}: many click events carry coordinates but no element name. We resolve them against the snapshot's recorded element tree: exact containment first (smallest containing element wins), then a $\pm$40\,px tolerance ring, then a coarse screen-zone fallback; every resolution is tagged \texttt{exact}, \texttt{tolerance}, or \texttt{zone}, and unresolvable clicks stay unresolved rather than being guessed. Third, \textbf{keyboard-layout mismatch}: some capture stacks record physical key positions decoded as QWERTY while the user types another layout. An explicit, operator-supplied translation map repairs this; it is identity by default and never inferred.

\subsection{Entity Typing}
Site parsers map URLs to the typed references of Section~\ref{sec:schema}: path and query parsing only, no fetching, no models. Resolution proceeds in layers: a bespoke parser for the host (the reference implementation ships more than twenty, covering professional networks, code hosting, search, documents, mail, maps, video, social, events, dashboards, AI chat, and local development), then a generic search-parameter detector, then a subdomain-and-path heuristic that types common infrastructure pages (sign-in, dashboard, email, calendar, meeting) even for sites without a bespoke parser, and finally a total fallback that guarantees every URL maps to something. Because every layer is a pure function of the URL, adding coverage is a contribution reviewable line by line.

\section{Reference Implementation}\label{sec:impl}
The reference implementation \cite{activityframes2026} is a Python package (MIT license) with three parts. The \textbf{capture engine} is provisioned on demand by \texttt{aframes record}: a pinned, MIT-licensed, open-source build that records application focus, window titles, URLs, the accessibility tree, and input events into a local SQLite database, entirely on-device, with audio capture off by default. Operators already running a compatible recorder can skip it and point the compiler at any existing database. The \textbf{compiler} has zero runtime dependencies, opens the database read-only, and exposes the document builders plus emitters for JSON, YAML, Markdown, and a compact plaintext context block designed for system prompts. The \textbf{MCP server} \cite{mcp2024} is a hand-rolled stdio JSON-RPC implementation, also dependency-free, exposing the six tools of Table~\ref{tab:mcp} to any MCP client. A workflow-pattern detector (repeated clicks, URL loops, action sequences, application-switching habits, daily habits) rounds out the surface.

\begin{table}[t]
\centering
\caption{MCP tool surface of the reference implementation.}
\label{tab:mcp}
\scriptsize
\begin{tabular}{p{2.4cm}p{4.9cm}}
\toprule
\textbf{Tool} & \textbf{Returns} \\
\midrule
\texttt{get\_context} & Compact chronological context block for the last $N$ hours, sized for inclusion in a system prompt \\
\texttt{get\_activity} & Full schema-v1 document (JSON) for a day or window \\
\texttt{get\_day\_summary} & Coverage plus per-app ledger (minutes, sessions, longest session) \\
\texttt{get\_patterns} & Repetitive workflows over the last $N$ days \\
\texttt{get\_communications} & Email/messaging surfaces with the window titles seen on each (measured tier; titles only) \\
\texttt{get\_steps} & Ordered click-by-click script behind one activity frame, for replay \\
\bottomrule
\end{tabular}
\end{table}

\section{Empirical Characterization}\label{sec:eval}
We characterize the system on the author's own live corpus: 61 calendar days of capture (46 days with activity) comprising 109{,}735 snapshot rows, 214{,}360 input events, and 8.4M element-tree rows across 54 applications (Table~\ref{tab:corpus}). The corpus is frozen: the recorder was migrated to a new database after the last captured day, so every number below is reproducible against an immutable file. The paper reads from this corpus at two freezes, labeled where used: this systems half uses the 2026-07-10 freeze (61 calendar / 46 active days, 109{,}735 frames), while the overhead measurements of Section~\ref{sec:ror} use a later 2026-07-22 freeze (51 active days, 128{,}756 frames). This is a single-user corpus; we report it as a characterization of the mechanism, not a user study (Section~\ref{sec:limits}).

\begin{table}[t]
\centering
\caption{Live capture corpus used throughout Section~\ref{sec:eval}.}
\label{tab:corpus}
\footnotesize
\begin{tabular}{lr}
\toprule
Calendar span & 61 days \\
Days with activity & 46 \\
Snapshot rows (frames) & 109{,}735 \\
\quad with URL & 59{,}458 \\
\quad idle-heartbeat rows & 20{,}429 (19\%) \\
Input events & 214{,}360 \\
Element-tree rows & 8{,}395{,}885 \\
Median intra-monitor gap & 6.1\,s \\
90th-percentile gap & 30.7\,s \\
Distinct applications & 54 \\
\bottomrule
\end{tabular}
\end{table}

\subsection{Token Cost}
\begin{figure}[t]
    \centering
    \begin{tikzpicture}
    \begin{axis}[
        ybar, ymode=log, log origin=infty,
        width=\linewidth, height=4.6cm,
        ylabel={tokens (log)}, ylabel near ticks,
        symbolic x coords={raw rows, frames JSON, context block},
        xtick=data, tick label style={font=\scriptsize},
        label style={font=\scriptsize},
        ymin=500, ymax=300000,
        bar width=17pt,
        nodes near coords, every node near coord/.append style={font=\scriptsize},
        point meta=explicit symbolic,
    ]
    \addplot[fill=measuredblue!35, draw=measuredblue] coordinates {
        (raw rows, 126812) [126{,}812]
        (frames JSON, 34815) [34{,}815]
        (context block, 1469) [1{,}469]
    };
    \end{axis}
    \end{tikzpicture}
    \caption{Token cost of one full day under three representations (cl100k\_base). Deterministic compilation yields an 86$\times$ reduction for prompt-ready context, at zero inference cost.}
    \label{fig:tokens}
\end{figure}
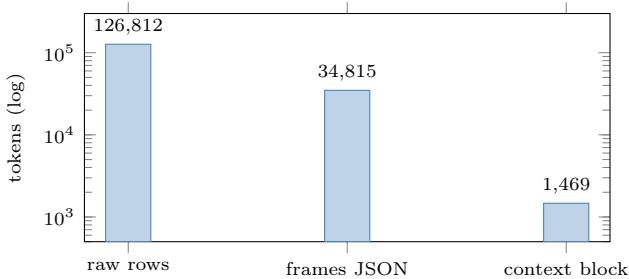

For one representative full day (2{,}066 snapshot rows), we compare three representations an agent could receive, tokenized with the \texttt{cl100k\_base} encoding. The raw rows, serialized as the JSON a search API would return, cost 126{,}812 tokens. The compiled schema-v1 document (222 frames at a 0.5-minute floor) costs 34{,}815 tokens, a 3.6$\times$ reduction that also relieves the agent of segmentation work. The compact context block costs 1{,}469 tokens, an 86$\times$ reduction, small enough to include in every system prompt (Figure~\ref{fig:tokens}). Both compiled forms are produced without any model call, so the reduction is free, and long-context results suggest the smaller representation is not merely cheaper but better used by the model \cite{liu2023lost}.

\subsection{Downstream Question Answering}\label{sec:qa}
Token cost is a means; the end is whether an agent can \emph{answer questions about the user's day}. We test this directly. We evaluate on eight days chosen by a fixed rule: the seven most recent consecutive active days plus the nearest preceding day whose raw serialization overflows the model's context window (included to exercise that regime). For each day, ground-truth answers are computed by an \emph{independent} SQL oracle over the raw tables: a documented inactivity-timeout dwell (credit each frame the gap to the next, capped at $60$\,s), deliberately not the activity-frames compiler, so the reference is not circular. Oracle and compiler measure the same construct, screen tenure: the capture heartbeat (Section~\ref{sec:compile}) keeps input-free stretches credited, so ``active minutes'' throughout this section means time a context was frontmost on an awake display, not interaction time. Every representation is graded against that one construct, which keeps the comparison fair, but absolute magnitudes should be read as tenure. We validate the compiler against the oracle before using the oracle to grade anyone. On the seven benchmark days whose capture was complete when the oracle was frozen, the compiler's covered active minutes agree with the oracle's total dwell to a median of $0.9$ minutes (all within $2.3$); the eighth day's capture continued after the freeze, so it is compared only at its snapshot. Distinct-application counts agree exactly or within one on every day. Two caveats keep this agreement honest. First, it compares aggregates that both approximate covered wall time, so it validates coverage rather than per-application arithmetic. Second, the two systems intentionally differ on multi-monitor days: the compiler credits each monitor's stream (Section~\ref{sec:compile}), while the oracle interleaves all monitors into one. On the context-overflow day, which has $133$ dual-monitor minutes, the dominant application earns $435.8$ minutes by the compiler's ledger but $346.1$ by the oracle, a $26\%$ divergence that the grading tolerance below happens to contain; switching the oracle's cap from $60$ to $90$\,s moves its day total by under $6$ minutes, so the divergence is the monitor convention, not the dwell cap. We report this rather than average over it. The generator emits 64 questions across five categories: which application dominated, how many active minutes in it, how many distinct applications, pairwise time ranking, whether a given domain was visited, and starting time, plus 16 \emph{absent-fact probes} (``did the user open Photoshop / visit netflix.com?'') that a faithful system answers negatively. We deliberately exclude two further question types. Longest-session length depends on the session-gap convention rather than a fact, and per-profile recall exceeds the compact block's token budget on busy days; each would penalize the compact block for a convention or a compression choice rather than for faithfulness, so we record the exclusion here to keep it on the books.

Two agents at different capability tiers, Claude Sonnet 4.5 and the larger Opus 4.5, each given no tools and answering only from the supplied text, answer every question from each of three representations: the raw rows, an LLM-generated summary of those rows written by that same agent, and the activity-frames output (the deterministic per-application ledger plus the compact context block, together $\approx$2k tokens). We note plainly that this hands the activity-frames agent measured per-application durations the other two must derive for themselves: that is exactly what the compiler is for, but it means the quantitative questions test whether the deterministic arithmetic was already done, not only whether the agent can read. Answers are graded against the oracle with fixed tolerances (numeric within $30\%$; times within $45$ minutes), and the ranking does not depend on them: at the Sonnet tier, a strict $10\%$/$15$-minute band leaves the three representations at $95.3\%$, $80.4\%$, and $66.1\%$, and the summary is so far off that widening the band to $50\%$ lifts it only to $67.9\%$. Table~\ref{tab:qa} reports both tiers at the default tolerance.

\begin{table}[t]
\centering
\caption{Downstream QA across two model tiers (8 days, 64 ground-truth questions, graded against the independent SQL oracle). ``Acc.'' is overall accuracy; ``Dur.\ err.'' is the mean absolute error on the dominant application's active minutes. The compiled block scores identically at both tiers, so a mid-tier model reads it as well as a frontier one; the raw-row and summary baselines improve with model strength but never catch it. Only the block is deterministic across runs and fits the context window on every day; on the busiest day the raw rows are $257$k tokens, so both baselines are infeasible and report on $7$ of $8$ days (the block, on all $8$).}
\label{tab:qa}
\footnotesize
\setlength{\tabcolsep}{4pt}
\begin{tabular}{lcccc}
\toprule
 & \multicolumn{2}{c}{Sonnet 4.5} & \multicolumn{2}{c}{Opus 4.5} \\
\cmidrule(lr){2-3} \cmidrule(lr){4-5}
Representation & Acc. & Dur.\ err. & Acc. & Dur.\ err. \\
\midrule
Raw rows & 82.1\% & 39.7\% & 91.1\% & 25.7\% \\
LLM summary & 66.1\% & 135.7\% & 80.4\% & 25.2\% \\
\midrule
\textbf{Activity frames} & \textbf{98.4\%} & \textbf{7.3\%} & \textbf{98.4\%} & \textbf{7.3\%} \\
\bottomrule
\end{tabular}
\end{table}

Four findings. First, \textbf{the gap is quantitative, not categorical.} At both tiers all three representations score perfectly on the absent-fact probes: the models do not invent applications or websites (hallucination rate $0\%$ throughout). The summary fails instead on \emph{magnitudes}. At the Sonnet tier it answers time questions at $7.1\%$ accuracy and mis-states the dominant application's active minutes by $135.7\%$ on average, against $7.3\%$ for activity frames (that residual is dwell-method rounding, not error). Fluent prose hides the damage: on 2026-07-05 the measured record (the oracle, with the compiler's ledger agreeing) is Google Chrome first at $161.6$ dwell minutes and Cursor second at $143.9$, over a day that ran $11{:}40$ to $22{:}26$, yet a Sonnet summary of that day names Cursor the primary application at ``$\sim$7 hours,'' inflating its $144$ measured minutes by $2.9\times$ and fabricating a nocturnal ``$6{:}40$\,PM--$5{:}26$\,AM'' session that spills into the next day. The activity-frames agent answered $160$ minutes. Every day, summary, and graded answer is in the released harness output.

Second, \textbf{a stronger model narrows the gap but does not close it, and the compiled block erases the difference between the two.} The frontier Opus model reconstructs durations far better from prose: its summary reaches $80.4\%$ overall and cuts the duration error to $25.2\%$. But that is still $3.5\times$ the block's $7.3\%$, and reading the block the two models are indistinguishable ($98.4\%$ each), whereas reading raw rows or a summary the frontier model runs $9$ to $14$ points ahead of the mid-tier one. The block lets the cheaper model answer as well as the expensive one; the baselines make capability matter. The benefit of deterministic compilation is therefore largest exactly where compute is cheapest to deploy.

Third, \textbf{the summary is non-deterministic}: regenerating it three times per day produced three distinct texts on every day at both tiers, whereas the activity-frames block was byte-identical across three regenerations, so the summary's errors are not even stable enough to correct for. Fourth, \textbf{the baselines do not always run.} The busiest day serializes to $257$k raw tokens, exceeding the context window, so both raw-row and summary consumption are infeasible; the $\approx$2k block is unaffected, which is why it alone reports on all eight days. Activity frames are not flawless: their one miss ($98.4\%$, not $100$, at both tiers) is a domain-recall question on the busiest evaluated day, where the compact block's budget drops a visited domain; that is the expected price of a bounded representation, and a recall question rather than a magnitude one. These accuracies are a single answering pass per representation per tier, at provider default sampling (model snapshots \texttt{claude-sonnet-4-5} and \texttt{claude-opus-4-5}, temperature and seed unpinned, so the non-deterministic baselines vary across re-runs); we release the full harness (oracle, question generator, and grader) so they can be repeated, with confidence intervals and further models, on any corpus.

\subsection{Latency and Reproducibility}
Compiling the same full day end to end (segmentation, enrichment-backed input accounting, entity typing, emission) takes a median of 68\,ms over five runs on a consumer laptop (Apple Silicon), with a 65 to 72\,ms range. Episodic memory at this cost can simply be rebuilt on every query. Reproducibility holds by construction and we verify it empirically: two independent compilations of the same window produce byte-identical documents once the generation timestamp is excluded. Determinism is what makes the memory cacheable, diffable across code versions, and testable in continuous integration.

\subsection{Entity Typing Coverage}
Across all 5{,}120 distinct URLs in the corpus, the layered parsers produce a non-generic typed reference for 81.3\%, spanning 46 kinds; the remaining 18.7\% fall back to the generic page reference with domain. The most frequent non-generic kinds are \texttt{profile} (1{,}106 distinct URLs), \texttt{search} (397), \texttt{email} (278), \texttt{event} (209), and \texttt{messaging} (200), reflecting that typed coverage concentrates precisely on the high-signal pages an agent most benefits from resolving. Coverage grows one pure function per site, so the long tail is closed incrementally by contribution rather than by any learned component.

\subsection{How Fragmented Is a Day?}
\begin{figure}[t]
    \centering
    \begin{tikzpicture}
    \begin{axis}[
        ybar, width=\linewidth, height=4.6cm,
        ylabel={frames}, ylabel near ticks,
        symbolic x coords={0-1, 1-2, 2-5, 5-10, 10-20, 20-45, 45+},
        xlabel={active duration (min)},
        xtick=data, tick label style={font=\scriptsize},
        label style={font=\scriptsize},
        ymin=0, ymax=13500,
        bar width=13pt,
        nodes near coords, every node near coord/.append style={font=\tiny},
    ]
    \addplot[fill=aigreen!30, draw=aigreen!70!black] coordinates {
        (0-1, 11842) (1-2, 2730) (2-5, 1952) (5-10, 641)
        (10-20, 275) (20-45, 70) (45+, 4)
    };
    \end{axis}
    \end{tikzpicture}
    \caption{Active-duration distribution of all 17{,}514 frames across 43 days (per-monitor segmentation). Median 0.5\,min, but 52\% of the sub-minute bar is single-snapshot transits rather than attention episodes; excluding them the median is 0.9\,min (see text).}
    \label{fig:durations}
\end{figure}
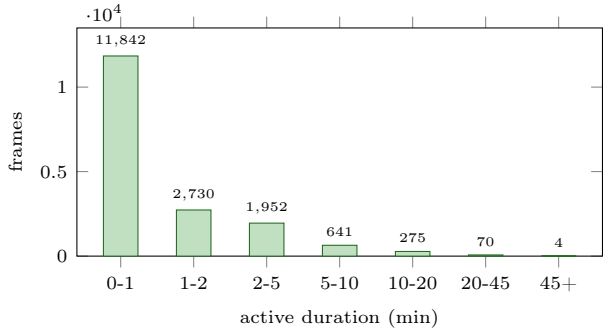

Compiling the 43 days that produce compiled frames (of the 46 with any capture in Table~\ref{tab:corpus}; the other three carry only idle-heartbeat rows) yields 17{,}514 frames (median 361 per day) with a median active duration of 0.5 minutes: 68\% of frames last under one minute (Figure~\ref{fig:durations}). A number this stark demands decomposition before interpretation, because two mechanical effects sit inside it. First, 52\% of the sub-minute frames (6{,}177) contain a single snapshot, with a median credit of 6.1 seconds: these are \emph{transits}, the application an operator passes through on the way somewhere else, faithfully recorded but not attention episodes. Second, a further 20\% (2{,}361) lie inside dual-monitor stretches, where each monitor earns its own stream by construction; 27\% of captured minutes on this corpus have two monitors active. Restricting to frames with at least two snapshots, the median rises to 0.9 minutes and 52\% remain under one minute; that residue is genuine switching. Read this way, the distribution is consistent with what interruption science has measured with dedicated instrumentation: three-minute working spheres \cite{gonzalez2004constant} and 75\% of laptop content segments under one minute \cite{yeykelis2014multitasking}. We deliberately do not present the raw histogram as an independent replication: an earlier draft of this analysis segmented all monitors as one interleaved stream, which shreds dual-monitor stretches and pushed the median to 0.3 minutes, and strict segmentation always overstates fragmentation until transits are separated out. The lesson is the schema's, not just ours: consumers get a \texttt{min\_minutes} floor (frames below it are omitted with the omission disclosed, never silently merged), the flicker-merge rule keeps sub-20-second detours from shredding genuine focus blocks while still recording them, and single-snapshot transits remain visible in the document precisely so that no compiler-side heuristic has to decide what counts as attention.

\section{The Routine Overhead Ratio}\label{sec:ror}
The compiler of Sections~\ref{sec:schema}--\ref{sec:impl} was built to produce episodic memory, but it is also an \emph{instrument}. Because it reduces a recurring stretch of capture to a short, replayable script by deterministic code alone, it lets us put a measured number on a quantity that agent-cost models assume but none of them observes: how much a computer-use agent overpays to re-derive a routine it has already done. We call that number the Routine Overhead Ratio $R$, and we report it, together with the recurrence $h$ (Section~\ref{sec:recurrence}), as the first readings of the two parameters those cost models are written in. Neither $R$ nor $h$ is offered as a principle; each is a measurement, taken with an open tool on one user's real work. The figures in this half are read from the same single-user author corpus characterized in Section~\ref{sec:eval}, at a later freeze: Section~\ref{sec:eval}'s systems numbers use the 2026-07-10 freeze ($61$ days, $46$ active, $109{,}735$ frames), while the overhead and reproducibility numbers here use the freeze extended through 2026-07-22 ($51$ active days, $128{,}756$ frames, $232{,}898$ input events). Between the two freezes the accessibility-tree coverage rose from about $41\%$ to $81.5\%$ of frames, which strengthens the structured grounding available to replay.

\subsection{Definition}
A \emph{routine} is a frequent action $n$-gram: a contiguous sequence of $k$ UI actions ($3 \le k \le 60$) that recurs at least three times within the corpus, restricted to sequences that name at least two distinct targets. Sessions are cut at inter-action gaps over $90$\,s. The two-named-target restriction is declared in advance, not tuned after the fact: without it, degenerate single-symbol repeats (a held key, a scroll loop) would satisfy ``recurs'' trivially and inflate every ratio, the granularity trap that any repetition metric on raw input invites. It is the same specificity that separates the delegable rate $h$ from the raw rate in Section~\ref{sec:recurrence}.

For a routine of $k$ steps we compare two token costs. The numerator is \textbf{modeled}. A memoryless, screenshot-driven agent re-derives the routine one step at a time, and per step consumes one screenshot, a fixed context read, and a fixed reasoning write:
\begin{equation}
C_\text{agent}(k) \;=\; k\left( \frac{w\,h}{750} \;+\; 350 \;+\; 180 \right),
\label{eq:num}
\end{equation}
the three per-step terms being the screenshot priced by Anthropic's $wh/750$ image-token rule for a $w\times h$ capture, a $350$-token context read, and $180$ tokens of reasoning output. At a typical $1512\times982$ capture this is about $2{,}500$ tokens per step ($1{,}979$ image $+\,350\,+\,180$), dominated by the screenshot. We are explicit that $C_\text{agent}$ is what published screenshot-driven agent loops \emph{would} spend, computed without executing an agent, and that it is an \emph{upper bound}: it assumes a screenshot baseline with no cross-step prompt caching, so an agent that reuses context across steps or grounds on the accessibility tree instead of a fresh screenshot spends fewer tokens per step and $R$ narrows accordingly. Section~\ref{sec:ror-limits} reports a live in-loop comparison against an accessibility-tree agent that saved only ${\approx}14\%$, and reserves the full live three-arm billing. The denominator is \textbf{measured}, and we report it as a \emph{ladder} rather than one figure, because the compiler can emit the routine at more than one level of detail, each a legitimate reading of $C_\text{replay}(k)$:
\begin{equation}
R = \frac{C_\text{agent}(k)}{C_\text{replay}(k)}, \quad C_\text{replay}(k) = \bigl|\texttt{tiktoken}(\cdot)\bigr|.
\label{eq:R}
\end{equation}
The compiler emits both rungs deterministically, with no model in the loop (\texttt{compile\_replay.py}, released with the reference implementation \cite{activityframes2026}); tokens are counted with \texttt{tiktoken} under \texttt{cl100k\_base}, the paper's encoding. The \emph{operational} rung is the full \emph{guarded skill plan}: a per-step sequence of typed actions, each carrying an expected-element, expected-role, and expected-application guard so replay can fail safe. Across the $20$ most frequent action routines these plans have a median of $247.5$ tokens and compile in a median of $0.5$\,ms at zero token cost, the compiler being CPU-only. The \emph{ceiling} rung is the \emph{minimal replay script} that strips the guards to the routine's bare information content: a median of $40$ tokens for a median $5$-step routine, about $8$ tokens per step. The guarded plan is roughly six times larger, which is precisely why it yields the smaller, more defensible ratio we lead with.

\subsection{The Ladder, and Its Distribution}
We lead with the \emph{conservative operational} number and keep the larger one as a ceiling, because leading with the ceiling would invite a cherry-picking objection. Against the guarded skill plan the compiler actually emits, the median is $R_\text{inject} = 60\times$ (interquartile $59$--$62$ across the $20$ compiled plans): injecting the compiled routine costs about one-sixtieth of re-deriving it from screenshots. Against the minimal script, the median reaches the information-content ceiling $R_\text{info} = 343\times$ at action granularity, the largest ratio the routine's bare description length can justify. Table~\ref{tab:ror} reports both rungs and, for $R_\text{info}$, the full distribution over routines. The ceiling is robust to the two knobs most likely to be accused of driving it: it rises monotonically but mildly with routine length (median $301\times$ at $k{=}3$--$4$, $358\times$ at $k{=}5$--$8$, $405\times$ at $k{=}9$--$15$, $401\times$ at $k{\ge}16$), and re-pricing the numerator at three screen configurations moves it only within a $259\times$ (conservative $1280\times800$) to $425\times$ (retina $1728\times1117$) band, so it is an artifact of neither short routines nor one display. At URL granularity the ceiling median is $198\times$. The most frequent action routine (a three-step compose-message loop, $181$ occurrences) and the most frequent URL routine (a mail-calendar-mail triangle, $25$ occurrences) are the mundane recurrences one would expect of real desk work, not exotic macros. We note the concentration plainly: the $20$ action routines behind $R_\text{inject}$ are dominated by two families, a compose-message loop and a window-close loop, sliced at several lengths by the $n$-gram miner, so $R_\text{inject}$ characterizes this user's highest-frequency micro-routines rather than a diverse cross-application task set; Section~\ref{sec:ror-limits} bounds the generalization.

\begin{table}[t]
\centering
\caption{The denominator ladder for $R = C_\text{agent}/C_\text{replay}$ (numerator modeled; denominators measured with \texttt{cl100k\_base}). $R_\text{inject}$ (operational) uses the guarded skill plan the compiler emits; $R_\text{info}$ (ceiling) uses the minimal replay script. $R_\text{inject}$ is computed on the $20$ most frequent action routines, so it carries no URL column. $R_\text{info}$ medians are priced at a typical $1512\times982$ screen; the resolution band re-prices the numerator at a conservative and a retina capture.}
\label{tab:ror}
\footnotesize
\setlength{\tabcolsep}{4pt}
\begin{tabular}{lcc}
\toprule
 & Action & URL \\
\midrule
Recurring routines & 614 & 261 \\
Routine instances & 5{,}508 & 1{,}303 \\
\midrule
\textbf{$R_\text{inject}$ (guarded plan, operational)} & \textbf{60$\times$} & n/a \\
\quad IQR / median plan tokens & 59--62$\times$ / 248 & n/a \\
\quad Compile cost $B$ & \multicolumn{2}{c}{$0.5$\,ms, $0$ tokens} \\
\midrule
$R_\text{info}$ (minimal script, ceiling) & 343$\times$ & 198$\times$ \\
\quad Interquartile range & 297--390$\times$ & 165--228$\times$ \\
\quad Recurrence-weighted mean & 337$\times$ & 194$\times$ \\
\quad Min--max & 167--509$\times$ & 67--301$\times$ \\
\quad Across screen resolutions & 259--425$\times$ & 150--245$\times$ \\
\quad Median script tokens & 40 & 43 \\
\quad Median steps $k$ / max occ. & 5 / 181 & 3 / 25 \\
\bottomrule
\end{tabular}
\end{table}

\subsection{Recovery, and a Modeled Three-Arm Comparison}
The recovery a hit yields depends on which rung of the ladder is used, and we attach the largest claim to the narrowest case. On a guard-matched step, deterministic \emph{local} replay --- \emph{parametric routine replay}: the routine's compiled structure replays deterministically while its variable slots carry the request's new values, so the agent picks the routine and fills the slots but never re-derives the steps --- takes the model out of the loop entirely, so it recovers
\begin{equation}
1 - \frac{1}{R} \;\approx\; 99\%
\label{eq:recovery}
\end{equation}
of that step's re-derivation cost (exactly $99.7\%$ at the median ceiling $R_\text{info}=343$). That $99\%$ is a per-covered-step ceiling, not a fleet saving: local replay reaches it only on the fraction of steps the compiler can guard, whose measured median is $0.415$, and it is exposed to interface drift that a live run must confirm (Section~\ref{sec:ror-limits}).

A modeled three-arm comparison makes the gap between ceiling and realized saving concrete. We price three ways of executing the $20$ recurring routines at Sonnet-class list rates (\$3/\$15 per Mtok, output fraction $0.07$), \emph{modeled from measured artifacts and token counts, not billed}: arm~A, a memoryless agent that re-derives each routine; arm~B, an agent given the compiled plan once and then acting with it in context; and arm~C, deterministic local replay. Relative to arm~A ($32.8$\,Mtok, \$125.81 over the fleet), injecting the plan (arm~B) saves $83.3\%$ of tokens (\$20.96, a $6.0\times$ per-occurrence reduction), and local replay (arm~C) saves $40.8\%$ (\$74.46, $1.7\times$). Two honesty notes. Arm~B's modeled saving assumes the screenshot baseline of Eq.~\ref{eq:num}; the one live in-loop comparison we have, against an accessibility-tree agent, saved only ${\approx}14\%$ (Section~\ref{sec:ror-limits}), so read $83.3\%$ as a modeled upper bound, not a measured in-loop figure. Arm~C falls well short of the $99\%$ ceiling for a structural reason: it credits only the guard-matched fraction, so the complementary $1-0.415$ deopt fraction still pays full arm-A price. The live-billed version of this table, with real usage JSON, is reserved (Section~\ref{sec:ror-limits}); we present it as modeled.

Across \emph{all} action steps rather than the recurring subset, the reachable saving is smaller still, because only a fraction $h$ of steps sit in a delegable routine. The honest all-fleet ceiling is $h\,(1-1/R_\text{info})$: with the in-sample delegable rate $h = 9.0\%$ it is $\approx 9.0\%$, and with the conservative out-of-sample rate $h = 7.7\%$ (Section~\ref{sec:recurrence}) it is $\approx 7.7\%$. Three disciplines keep every figure here honest. First, no rung of $R$ is ever multiplied by the $86\times$ context compression of Section~\ref{sec:eval}: those are different tokens (reading a day versus re-deriving a routine), and stacking them would double-count the same work. Second, $R$ is never combined with prompt-cache or KV-cache discounts; caching and replay are alternative recoveries of the same repetition, not additive ones. Third, both the numerator and the three-arm dollars are modeled, so we report them as ceilings and reserve the live billing.

\subsection{Desktop Routine Recurrence $h$}\label{sec:recurrence}
The ratio $R$ prices a single hit; the recurrence $h$ is how often hits occur, measured as the fraction of action steps that fall inside a recurring routine. A single number would mislead, so we report $h$ at two levels. The \emph{raw} rate is $h_\text{raw} = 83.1\%$: the fraction of action steps inside \emph{any} recurring $n$-gram. This is dominated by the generic micro-structure of input, the type-a-character, return-to-field, type-again texture that repeats constantly and carries no reusable work. The \emph{delegable} rate is $h_\text{specific} = 9.0\%$ at action granularity ($13.1\%$ at URL granularity): the fraction of steps inside the $\ge 2$-named-target routines of Section~\ref{sec:ror}, that is, inside an identifiable, repeatable task. We report the gap $83.1\% \rightarrow 9.0\%$ openly rather than quoting the larger figure: it is the difference between ``the keyboard repeats'' and ``the work repeats,'' and only the latter is a candidate for delegation.

\paragraph{A temporal holdout (the single predictive claim)} A within-sample recurrence rate is partly circular, since a routine is counted as recurring in part because we already watched it recur. To obtain a non-circular value we fit the routine table on the first $40$ active days ($4{,}847$ routine signatures) and then ask what fraction of action steps on the held-out final $11$ days fall into a routine \emph{already known} from the training window. The out-of-sample predicted hit rate is $7.7\%$, against an in-sample $8.6\%$ on the training days themselves. The modest $8.6\% \rightarrow 7.7\%$ drop is the within-user temporal-drift gap (a same-user, later-in-time split, not a cross-population holdout), and $7.7\%$ is the recurrence the cost accounting should carry; the all-fleet ceiling $h\,(1-1/R_\text{info}) \approx 7.7\%$ follows from it. We make exactly one predictive claim, and this is it.

\paragraph{What $h$ is not} Web-era studies of page revisitation report far larger constants, on the order of $40$ to $58\%$ of page visits being revisits \cite{tauscher1997revisit, adar2008revisit}. We do not claim those as $h$. A page revisit is not a delegable task: back-button noise, re-checking a feed, and reopening a tab are revisits with no reusable work content, and importing that number would overstate $h$ by roughly an order of magnitude. Our $h$ measures recurring \emph{task} structure in real desktop work, and on this corpus that quantity is $\approx 0.08$--$0.13$, not $\approx 0.5$. Reporting the smaller, honest figure is the point of the specificity rule.

\subsection{Reproducibility as a Certified Property}\label{sec:repro-cert}
Because compilation contains no model, reproducibility is not a behavior we hope for but a property we can certify. A certification harness (\texttt{certify\_ivm.py}, released with the code \cite{activityframes2026}) checks three conditions over the $51$-active-day corpus and returns \textsc{pass}. First, \textbf{byte-identical output}: re-compiling any day twice yields byte-identical documents once the single emission-metadata field \texttt{generated\_at} (a wall-clock stamp of the run) is excluded. We name that exclusion rather than quietly canonicalizing it away; it is the one field that is not a function of the capture. Second, \textbf{rebuild equals incremental}: rebuilding the entire history from scratch produces the same bytes as compiling day-by-day, and earlier days are never rewritten by later capture (append-only). Third, \textbf{compile cost does not grow with history}: the median full-day compile across the run is $220.9$\,ms, and the mean over the second half of the corpus is $0.86\times$ the mean over the first half ($216.4$\,ms versus $251.8$\,ms), so per-day cost tracks the size of the day's delta, not the length of accumulated history; it is $O(|\Delta|)$. (The $68$\,ms of Section~\ref{sec:eval} is one representative lighter day; the certification spans light and heavy days, from $0.1$ to $930$\,ms.)

We claim exactly one thing from this, and concede its lineage. The claim is a CI-checkable byte-equality contract on a stateless projection of an append-only capture log. We claim \emph{no} novelty in incremental view maintenance: deterministic, incrementally maintainable views over append-only logs are a mature area, with DBSP giving automatic incremental view maintenance for rich query languages \cite{budiu2022dbsp}, and event-sourcing long rebuilding state as a fold over an immutable event log. Our contribution is not the mechanism but its \emph{use as a certified guarantee for agent memory}: an episodic-memory artifact whose equality across runs, across code versions, and across rebuild strategies is enforced by a test, which is what makes the memory safe to cache and mechanically auditable in the sense Section~\ref{sec:limits} requires.

\section{Privacy, Trust, and Limitations}\label{sec:limits}

\subsection{Privacy Model}
The entire pipeline is local: capture, storage, and compilation happen on the user's machine, the compiler opens the capture database read-only, and nothing is transmitted anywhere by the system itself. The operator chooses what leaves, and when, by handing a compiled artifact to an agent. The schema-level content rule (Section~\ref{sec:schema}) keeps typed text out of documents by default; audio capture is off by default in the provisioned engine. Compilation itself involves no model of any kind; the capture engine does run on-device OCR to read screen content, but that output never leaves the machine, and no language model, local or remote, participates in producing memory. The capture database itself remains sensitive at rest, as membership-inference work on memory stores reminds us \cite{chen2026mrmmia}; we treat device-level encryption as the appropriate control and note that this risk is shared by every capture system rather than introduced by compilation.

\subsection{Trust Properties}
Two structural properties address emerging attacks on agent memory. Evidence pointers make every episode mechanically auditable: a verifier can re-read the named raw rows and recompute the frame, which raises the bar for poisoning attacks that rely on unverifiable memories \cite{louck2026securing}. The measured/inferred boundary ensures that even a compromised or careless tier-2 tool cannot inject interpretation disguised as observation; consumers can always strip to the measured core. We do not claim these properties defeat a compromised capture engine, which can fabricate raw rows; provenance begins at the database.

\subsection{Limitations}
Four limitations bound our claims. First, the empirical characterization is a single-user corpus (one professional, one machine, 61 days); cadence, fragmentation, and entity coverage will differ across roles and platforms, and a multi-user study is future work. Second, the reference implementation reads one capture engine's database layout; the schema is engine-agnostic but each new source needs an adapter. Third, the measured tier is structurally silent on intent: it reports two profile views and a people search, never ``prospecting,'' and consumers who need intent must add tier-2 inference and accept its tags. Fourth, screen presence is an imperfect proxy for attention, and on this corpus the gap is quantifiable rather than hypothetical. The capture heartbeat keeps input-free stretches credited: intervals that terminate in a heartbeat row carry 45\% of all credited active time, and uninterrupted heartbeat runs reach 42 minutes, so dwell includes reading and watching but also time the user had stepped away from an awake display. The dwell cap does not bound this; it only bounds credit across capture stalls and display sleep. Frames report input volume precisely so consumers can gate on interaction, and a tier-2 tool may label presence-only stretches, but the measured tier reports tenure, not attention, and consumers must read it that way. Monitors compound the proxy error in the other direction: each records its own stream, so a minute with two active monitors earns credit twice in per-application ledgers (27\% of captured minutes here; Section~\ref{sec:qa}). Off-screen work (paper, phone, conversation) appears only as gaps. Relatedly, frames measure attention episodes, not tasks: interruption research shows interrupted tasks are resumed across much longer horizons than any single frame \cite{gonzalez2004constant}, so task-level structure belongs to tier-2 inference. Finally, the downstream benchmark (Section~\ref{sec:qa}) evaluates two agent tiers (Claude Sonnet 4.5 and Opus 4.5) on this single-user corpus with one answering pass each. A stronger model narrows the summary's gap, so the magnitude of the effect is not model-independent; but the block's determinism, its context-fit, and its being read equally well by both tiers are structural properties no model strength supplies, and we release the harness so the comparison can be rerun with confidence intervals, against other model families such as GPT and Gemini, and, as multi-user capture becomes available, other users.

\subsection{Limitations of the Overhead Measurements}\label{sec:ror-limits}
Five limitations bound the results of this half of the paper, beyond the systems limitations already stated in Section~\ref{sec:limits}.

\textbf{Single user.} Every number in Section~\ref{sec:ror} comes from one user's machine, the author's, over $51$ active days ($128{,}756$ frames). We state this plainly: $R$ and $h$ are properties of this person's work, not of a population. The specificity rule and the temporal holdout guard against granularity and circularity artifacts, but not against having sampled one professional on one platform. A multi-user replication, and a public-dataset second subject, are the obvious next step and are not claimed here.

\textbf{The numerator and the three-arm dollars are modeled, not billed.} $R$'s numerator (Eq.~\ref{eq:num}) is what a memoryless screenshot-driven loop would spend under the Anthropic image-token rule and fixed per-step budgets, priced without running an agent, and the three-arm comparison of Section~\ref{sec:ror} is likewise modeled from measured artifacts (compiled plans, token counts, guard coverage) at list prices, not billed. The denominators themselves are real: the guarded plan and the minimal script are emitted and tokenized deterministically, and the median guard coverage of $0.415$ that bounds local replay (arm~C) is measured, not assumed. Since the freeze, one owner-authorized live execution has been run, and we report it as a first confirmation of the replay side. A compiled two-step routine (open a compose surface, type a draft; nothing was ever posted) was matched to a natural-language request by a small \emph{local} model (a Tier-2 retrieval step we do not count in the execution total) and then executed in a live, authenticated browser session by the released executor: both steps grounded by accessibility role+name, \emph{zero} model tokens \emph{at execution}, a few seconds of wall-clock. The element references differed between two runs of the same plan and name-based grounding adapted; on a wrong page the same plan grounded nothing and performed zero actions, the intended fail-safe. The accessibility snapshots an in-loop agent would read to choose each action measured ${\approx}10.5$k tokens per step on the same pages; a separate live comparison that kept the model \emph{in} the loop with the compiled plan as context saved only ${\approx}14\%$ against an accessibility-driven agent, so the large ratios of Section~\ref{sec:ror} require the model fully out of the loop, which parametric replay is. Three bounds on this confirmation: it is one two-step routine; its plan was seeded from live accessibility names standing in for a mined routine (click-level grounding of the recorder is still under validation); and the guard-miss deopt path was not exercised. The full live three-arm billing with real usage JSON remains reserved. Read $R$ as a modeled ratio whose denominator --- including its zero-token on-hit case --- is now live-confirmed, and the dollar savings as a modeled ceiling awaiting live billing.

\textbf{Capture is not free.} The instrument has an operating cost, which we report rather than hide. The corpus database is $9.5$\,GB, about $0.19$\,GB per active day; on-device OCR runs on $96\%$ of frames, a continuous duty cycle; and the accessibility tree that grounds replay is present on $81.5\%$ of frames, so $18.5\%$ offer no structured target and would fall back to coordinates. Replay coverage is therefore bounded by that $81.5\%$, and the compression and QA wins of Section~\ref{sec:eval} are gross of the capture engine's own footprint.

\textbf{Match precision propagates into $q$.} Turning a captured routine into a replayable script depends on entity typing, which is approximately $82\%$ accurate at assigning a non-generic type (Section~\ref{sec:eval}). Typing errors propagate directly into the match-precision term $q$ of Eq.~\ref{eq:amortized}: a mistyped target can match the wrong routine or fail to match a real one, so $q < 1$ and the honest fleet saving carries that factor. We do not assume $q = 1$.

\textbf{OCR is a model, so determinism holds forward, not back to pixels.} The only learned component anywhere in the pipeline is the on-device OCR that reads screen text at capture time. Byte-level reproducibility (Section~\ref{sec:repro-cert}) therefore holds from the \emph{stored OCR text} forward: given the captured text, every downstream document is a deterministic function of it. It does not hold back to the pixels, because a different OCR model, or a re-scan of the same screenshots, could yield different text. We scope the determinism claim to the compile path over stored capture, never to perception.

\section{Conclusion}
Agents are blind to the activity stream that most defines their user's day, not because capture is missing but because nothing turns capture into memory an agent can afford, reproduce, and trust. Activity frames fill that seam with the least interesting tool available, deterministic code, and we argue that this dullness is the point: at 68\,ms and zero tokens per day, episodic memory becomes infrastructure rather than inference, and the measured/inferred boundary gives interpretation a place to live without contaminating fact. The same deterministic compiler doubles as a demand-side instrument: because it reduces recurring capture to a replayable script by code alone, it reads the cost parameters $R$ and $h$ that agent-cost models assume, here as first single-user values, with a modeled numerator, awaiting the multi-user replication and live billing we reserve. The schema, compilation rules, and implementation are open \cite{activityframes2026}; we hope the format outlives the reference code, and that capture systems, memory layers, and agents converge on a shared, honest representation of what a person actually did.

\bibliographystyle{ieeetr}
\bibliography{ref,ref_breakthrough_fixed}

\end{document}